\documentclass[11pt,a4paper]{article}
\usepackage[hyperref]{emnlp2020}
\usepackage{latexsym}
\usepackage{textcomp}
\usepackage{makecell}
\usepackage{times}
\usepackage{url}
\usepackage{latexsym}
\usepackage{array}
\usepackage{pgfplots}
\usepackage{graphicx} 
\usepackage{subcaption}
\usepackage{xcolor}
\usepackage{blindtext}%
\usepackage{tabularx}
\usepackage{rotating}

\usepackage{sidecap}

\newcolumntype{Y}{>{\centering\arraybackslash}X}
\newcolumntype{s}{>{\hsize=0.69\hsize}X}

\graphicspath{ {./} }

\usepackage{microtype}

\aclfinalcopy % Uncomment this line for the final submission
\def\aclpaperid{***} %  Enter the acl Paper ID here

\title{Investigating African-American Vernacular English in Transformer-Based Text Generation}
 \author{Sophie Groenwold\thanks{\hspace{1 mm} Equal contribution.}, Lily Ou\footnotemark[1], Aesha Parekh\footnotemark[1], Samhita Honnavalli\footnotemark[1],\\ \textbf{Sharon Levy, Diba Mirza, William Yang Wang} \\
University of California, Santa Barbara \\
\texttt{ \{sophiegroenwold, lilyou, aeshaparekh, shonnavalli\}@ucsb.edu }
\\
\texttt{ \{sharonlevy, dimirza, william\}@cs.ucsb.edu }
}
\date{}
\begin{document}
\maketitle
\begin{abstract}
The growth of social media has encouraged the written use of African American Vernacular English (AAVE), which has traditionally been used only in oral contexts. However, NLP models have historically been developed using dominant English varieties, such as Standard American English (SAE), due to text corpora availability. We investigate the performance of GPT-2 on AAVE text by creating a dataset of intent-equivalent parallel AAVE/SAE tweet pairs, thereby isolating syntactic structure and AAVE- or SAE-specific language for each pair. We evaluate each sample and its GPT-2 generated text with pretrained sentiment classifiers and find that while AAVE text results in more classifications of negative sentiment than SAE, the use of GPT-2 generally increases occurrences of positive sentiment for both. Additionally, we conduct human evaluation of AAVE and SAE text generated with GPT-2 to compare contextual rigor and overall quality.
\end{abstract}

\section{Introduction}

African American Vernacular English (AAVE) is a sociolinguistic variety of American English distinct from Standard American English (SAE) with unique syntactic, semantic, and lexical patterns \cite{aave-introduction, jones_taylor}. Millions of people from predominately Black communities in the United States and Canada use variants of AAVE on a daily basis. Although AAVE has historically been used in spoken contexts, the growing use of social media has encouraged AAVE in written media for which NLP models are increasingly being used.

Recent work in Natural Language Generation (NLG) has introduced GPT-2, a Transformer-based language model that generates high-quality, coherent 
text when prompted by arbitrary input \cite{radford2019language}. However, GPT-2 displays bias towards particular social groups \cite{solaiman2019release}. \newcite{sheng-etal-2019-woman} shows that NLG tools are biased with regard to the subject of a sentence when that subject belongs to an underprivileged group, and \newcite{shen-et-al} tests sentiment analysis tools with intent-controlled pairs with varying stylistic inclinations. Studies regarding AAVE have analyzed tasks such as POS tagging \cite{jorgensen-etal-2016-learning}, detecting AAVE syntax \cite{stewart-2014-now}, voice recognition and transcription \cite{dorn-2019-dialect}, dependency parsing \cite{blodgett-etal-2016-demographic}, and hate speech detection \cite{sap-etal-2019-risk}, but not language generation. Coupled with concerns that NLG tools can be used for generating fake news \cite{gehrmann-etal-2019-gltr} or impersonating internet users \cite{NIPS2019_9106}, it is important that current work investigates the contexts in which NLG models display bias against certain demographics.

In this paper, we examine the bias of GPT-2 text generation against AAVE features. We create a new dataset of AAVE/SAE content-controlled pairs by retrieving AAVE tweets and employing human translators to obtain their SAE counterparts. By doing so, we isolate AAVE syntactic structures and lexical items. We then prompt GPT-2 with the first segments of each AAVE/SAE pair. The generated text is compared to its corresponding original second segment by BLEU, ROUGE, and sentiment scores. Additionally, we provide human evaluation for the generated text based on context and quality. 

\begin{figure*}
\center
    \includegraphics[height=4.5cm]{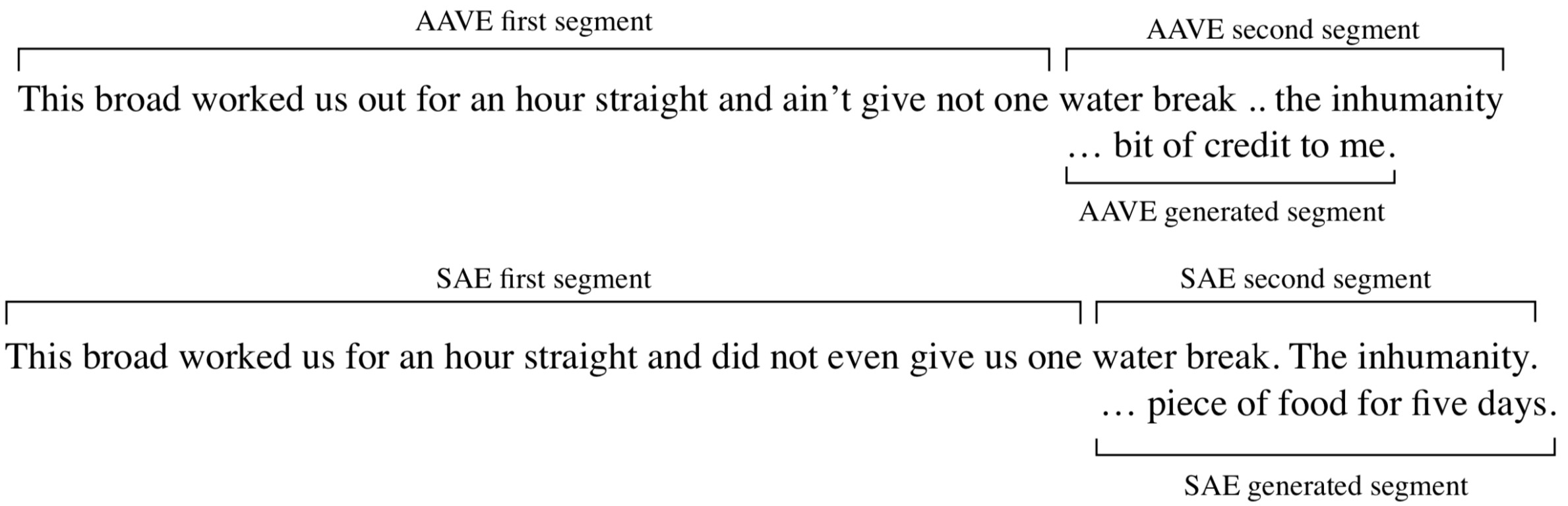}
    \caption{Terms used to refer to segments of each AAVE/SAE pairwise sample. Each first segment is used to prompt its respective generated segment and sentiments are taken of the second and generated segments.}
   \label{segment-example}
\end{figure*}

Thus, our contributions include:
\begin{itemize}
  \item[$\bullet$] An intent-equivalent dataset of AAVE/SAE pairs with differences only in syntactic structure and dialect-specific vocabulary.
  \item[$\bullet$] New evaluation of GPT-2 using sentiment analysis, BLEU, and ROUGE scores of its generated text and the original SAE and AAVE segments.
  \item[$\bullet$] Human evaluation of GPT-2 generated text for each AAVE/SAE pair, where evaluation is conducted to identify contextual accuracy, quality, and likelihood of being categorized as machine-generated. 
\end{itemize}

\section{Dataset}
Our dataset consists of tweets identified as having  at least 99.9\% confidence of using AAVE lexical items by the TwitterAAE dataset \cite{blodgett-etal-2016-demographic}. We then obtain the SAE equivalent of each of these tweets by employing Amazon Mechanical Turk (AMT) annotators for a total of $n = 2019$ AAVE/SAE pairs. The average length of the original AAVE tweets is about 21 words, and the average length of the SAE counterparts is about 22 words. These samples are intended to be used as a test set for probing neural language model-based text generation.

We use the terms ``first segment," ``second segment," and ``generated segment" to refer to the different sections of each AAVE/SAE sample throughout this paper. A visualization of these partitions can be seen in Figure \ref{segment-example}.

\paragraph{Sample Identification}

TwitterAAE \cite{blodgett-etal-2016-demographic} collects AAVE tweets by using a distantly supervised mixed-membership model on samples that are geolocated to African-American blockgroups, as defined by the U.S. Census data. The tweets have been filtered to ensure conversational language and verified as AAVE on the basis of AAVE-specific lexical item inclusion, phonological phenomena in orthographic variation, and syntactic construction. From TwitterAAE, we randomly sample tweets that contain at least 15 words and have a posterior probability of being demographically-aligned to AAVE of at least 99.9\%. We remove hashtags as they are social media-specific occurrences and emoji since we expect them to have disproportionate influence on sentiment scores.

\begin{table*}[t]
\small
\centering
\setlength{\tabcolsep}{5.5pt}
\begin{tabular}{l| cccc| cccc}
 & \multicolumn{4}{c}{\textbf{SAE}} & \multicolumn{4}{c}{\textbf{AAVE}}\\
\hline
 \textsc{DistilBERT Classifier} & \textit{positive} & \textit{negative} & \textit{neutral} & \textit{average}& \textit{positive} & \textit{negative} & \textit{neutral} & \textit{average}\\
Original Second Sentiment & 50.12\% & 49.88\% & N/A & 0.0066 & 42.35\% & 57.65\% & N/A & -0.1436 \\
GPT-2 Generated Sentiment & 47.75\% & 52.25\% & N/A & -0.0399 & 46.16\% & 53.84\% & N/A &-0.0769 \\
\hline
 \textsc{VADER \& TextBlob Average} & \textit{positive} & \textit{negative} & \textit{neutral} & \textit{average} & \textit{positive} & \textit{negative} & \textit{neutral} & \textit{average} \\
 Original Second Sentiment & 24.67\% & 20.28\% & 55.05\% & 0.0779 & 25.06\% & 19.71\% & 55.23\% & 0.0325 \\
GPT-2 Generated Sentiment & 66.02\% & 25.73\% & 8.25\% & 0.1909 & 62.43\% & 32.66\% & 4.85\% & 0.1443 \\

\end{tabular}
\caption{Sentiment scores and averages for the SAE and AAVE samples in our dataset, using pretrained DistilBERT, VADER, and TextBlob sentiment classifiers.}
\label{fig:avg-and-concentration-results}
\vspace{-3ex}
\end{table*}

\paragraph{Pairwise Sample Collection}

To investigate GPT-2 generated text on AAVE versus SAE, we use (small) GPT-2 \cite{radford2019language} from Open-AI for text generation, which is pretrained on out-bound sources from Reddit comments with at least three karma. 

Although prior work exists in using unsupervised word embeddings to create vector space-aligned demographic translations \cite{shen-et-al, lample-et-al}, we instead use human translation for accuracy purposes. We therefore employed AMT annotators to obtain the SAE equivalents of our AAVE samples.

Each AMT worker was given an AAVE tweet sample, first as a whole for context and then split into a first segment and a second segment. The latter consisted of the last five words of the sample, so as to take approximately a third of the full sample (see Figure \ref{segment-example}). We asked annotators to translate the first and second segments individually into SAE; this partition was necessary for use with GPT-2, BLEU, and ROUGE. We provided example translations, and the full instructions can be seen in the \nameref{subsec:aave-to-sae} annotation guidelines. Annotators were filtered by HIT approval rate (higher than 97\%) and location (within the United States). Additional instructions included either expanding or providing a contextual equivalent for acronyms, insertion of SAE-appropriate grammar, and preservation of overall structure and intent of the AAVE sample. Annotators were also told to translate the n-word, but to retain non-AAVE-specific explicit language. 

\paragraph{Dataset Viability}

We test the variability of our dataset’s results by taking 1000 random partitions of size 1500 and use DistilBERT \cite{sanh2019distilbert} to find the average sentiment score. For each partition of our data (both SAE and AAVE with and without generation by GPT-2), the sample variance is under 0.02\%. 

\paragraph{Semantic Evaluation}

Previous work has shown that non-AAVE speakers often fail to demonstrate comprehension of AAVE speech, and we acknowledge that such misunderstandings may influence the intent-equivalence of our dataset \cite{testifying-while-black}. Thus, to determine the semantic validity of the translations, we asked annotators who self-identified as native AAVE speakers and/or code-switchers to verify whether translated SAE phrases preserved the meaning of original AAVE phrases. Of 156 randomly sampled AAVE/SAE pairs, 90\% are intent-equivalent according to native AAVE speakers, and 95\% according to code-switchers.  This confirms that the majority of our pairs have semantic equivalence. We have included the instructions for this validity check in the \nameref{subsec:semantic-equivalence-guidelines}.

\section{Sentiment Analysis}

We use a sentiment analysis pipeline from Huggingface\footnote[1]{https://huggingface.co/Transformer/main\textunderscore classes/\linebreak pipelines.html}, to evaluate the sentiment of our samples. The pipeline uses distilbert-base-uncased-finetuned-sst-2-english\footnote[2]{https://huggingface.co/distilbert-base-

uncased-finetuned-sst-2-english}, which is pretrained on movie reviews from the Stanford Sentiment Treebank  \cite{socher-etal-2013-recursive}. In addition to the DistilBERT sentiment classifier, we use VADER, which is a lexicon and rule-based sentiment analysis tool that is attuned to social-media specific sentiment intensity  \cite{vader}, and TextBlob\footnote[3]{https://textblob.readthedocs.io/en/dev}, which does not have documentation on its implementation. However, we justify our use of the latter through its widespread use as an off-the-shelf sentiment classifier, such as in \newcite{sheng-etal-2019-woman}.

The DistilBERT sentiment classifier restricts classifications to either positive or negative, with degrees of confidence ranging from 0 to 1; we translate this to a -1 to 1 negative-to-positive scale. From VADER we use the compound score, and from TextBlob the polarity; both metrics are normalized and weighted and thus also range from -1 to 1. VADER and TextBlob scores include 0.0, or neutral, while the DistilBERT sentiment classifier does not. We average the latter two in Table \ref{fig:avg-and-concentration-results} to account for model variability in the sentiment classifiers, but keep the DistilBERT scores separate because it does not include neutral classifications.

\paragraph {Baseline}

As a baseline, we compare the sentiment of each AAVE original second segment to its respective SAE original second segment. We observe that the pretrained sentiment analysis models categorize AAVE as more negative than SAE, despite having the same intent. AAVE has 157 (7.7 \% percent) more negative instances than it does positive when using DistilBERT and 37 (1.8 \% percent) more negative and neutral instances when using the VADER-TextBlob average. The VADER-TextBlob averages appear to be less biased against AAVE than DistilBERT. 

\paragraph {Sentiment Comparison of Generated Text}

To determine whether GPT-2 generates more negative phrases when provided AAVE text, we compare the sentiment of the generated segment for AAVE to its corresponding generated segment for SAE. For DistilBERT we see that the average for AAVE generated segments is -0.0769, while its SAE counterpart is -0.0399 (see Table \ref{fig:avg-and-concentration-results}). This indicates that the AAVE GPT-2 generated segments are more negative than their corresponding SAE segments. We see the same trend for the VADER and TextBlob averages, where the AAVE generated segment has a more negative sentiment score than its corresponding SAE segment. Additionally, in the case of the VADER-TextBlob average, the negative sentiments of the original second segments for SAE and AAVE differ by a margin of 0.57\%, whereas the difference between the generated negative sentiments is 6.93\%, with AAVE being more negative. This shows that even though AAVE has more positive instances than SAE for its original second segment, the use of GPT-2 increases negative sentiment more for AAVE than for SAE.

We also perform a McNemar-Bowker significance test on the results from Table \ref{fig:avg-and-concentration-results} and find a significant difference between the original and generated sentiments for DistilBERT AAVE, VADER AAVE and SAE, and TextBlob AAVE and SAE with $\alpha = 0.05$. VADER and Textblob for both AAVE and SAE had $p < 0.01$. DistilBERT for AAVE had $p = 0.012$ and DistilBERT for SAE had $p = 0.11$.

\paragraph {Flipped Sentiment}

We compare the sentiment of the second segment of each AAVE phrase to the sentiment of its generated segment and do the same for each corresponding SAE sample. This allows us to observe the extent to which GPT-2 flips the sentiment from positive to negative and vice versa, and whether flipping from positive to negative sentiment is more prevalent in AAVE.

We find that AAVE samples have lower sentiment scores than their SAE equivalents with the classifiers we utilized. However, the AAVE generated segments increase in DistilBERT sentiment score going from -0.1436 to -0.0769 on the -1 to 1 scale, while SAE generated segments decrease from 0.0066 to -0.0399 (see Table \ref{fig:avg-and-concentration-results}). However, this is not the case with the VADER-TextBlob average, as the sentiment scores increase for both AAVE and SAE generated segments when compared to their respective second segments.

For the VADER-Textblob average in Table \ref{fig:avg-and-concentration-results}, AAVE generated segments are 50.38\% less neutral than their original second segments, and SAE generated segments are 46.8\% less neutral. While the majority of the original second segments are classified as neutral, the majority of the generated segments are instead classified as positive. However, SAE has a larger increase in positive sentiment scores than AAVE, even though its original positive sentiment was lower than AAVE's corresponding original sentiment. 

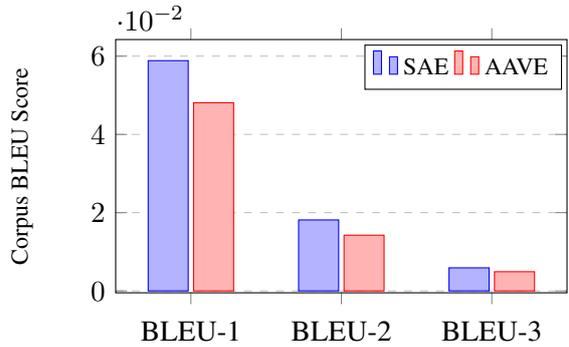
\begin{figure}[t!]
    \begin{minipage}{0.47\textwidth}
    \centering
        \begin{tikzpicture}
            \begin{axis}[
                width=1\textwidth,
                height=0.2\textheight,
                enlarge x limits=0.25,
                ybar,
                bar width=15pt,
                x label style={at={(axis description cs:0.5,-0.05)},anchor=north},
                ylabel={\small{Corpus BLEU Score}},
                symbolic x coords={BLEU-1, BLEU-2, BLEU-3},
                xtick = {BLEU-1, BLEU-2, BLEU-3},
                xticklabel style={text height=2ex},
                legend pos=north west,
                ymajorgrids=true,
                grid style=dashed,
                legend style={at={(0.77, 0.98)},
                  anchor=north,legend columns=-1},
            ]
            \addplot coordinates { 
                (BLEU-1,0.058849)
                (BLEU-2,0.018144)
                (BLEU-3,0.005906)
            };
            \addplot coordinates { 
                (BLEU-1,0.048096)
                (BLEU-2,0.014239)
                (BLEU-3,0.004908)
            };
            \legend{\small{SAE}, \small{AAVE}},
            \end{axis}
        \end{tikzpicture}
        \caption{BLEU scores for text generated by GPT-2.
        }
        \label{bleu-fig}
    \end{minipage}
\end{figure}

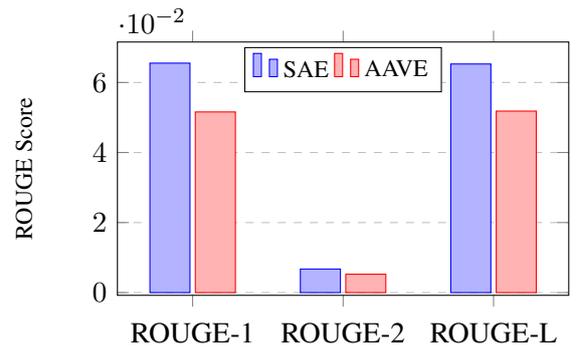
\begin{figure}[t!]
    \begin{minipage}{0.47\textwidth}
        \begin{tikzpicture}
            \begin{axis}[
                width=1\textwidth,
                height=0.2\textheight,
                enlarge x limits=0.25,
                ybar,
                bar width=15pt,
                x label style={at={(axis description cs:0.5,-0.05)},anchor=north},
                ylabel={\small{ROUGE Score}},
                symbolic x coords={ROUGE-1, ROUGE-2, ROUGE-L},
                xtick = {ROUGE-1, ROUGE-2, ROUGE-L},
                xticklabel style={text height=2ex},
                ymajorgrids=true,
                grid style=dashed,
                legend style={at={(0.5, 0.98)},
                  anchor= north,legend columns=-1},
            ]
            \addplot coordinates { 
                (ROUGE-1,0.06552536247)
                (ROUGE-2,0.006733220419)
                (ROUGE-L,0.06527926447)
            };
            \addplot coordinates { 
                (ROUGE-1,0.05158959854)
                (ROUGE-2,0.00526073719)
               (ROUGE-L,0.0518300273153914)
            };
            \legend{\small{SAE}, \small{AAVE}},
            \end{axis}
        \end{tikzpicture}
        \caption{ROUGE scores for text generated by GPT-2.}
        \label{rouge-fig}
    \end{minipage}
    \vspace{-2ex}
\end{figure}

\section{Quality of Generated Text}

We use BLEU, ROUGE, and human evaluation scores to determine the difference in the quality of GPT-2 generated text for SAE and AAVE samples.

\paragraph{BLEU and ROUGE}

For all SAE and AAVE samples, we isolate the second segment of the original sample, for which we take the last five words, and the first five words generated by GPT-2. We then compare the generated segment to the original second segment by calculating their BLEU and ROUGE scores. Specifically, ROUGE-1 and ROUGE-2 measure the overlap of unigrams and bigrams respectively, and ROUGE-L identifies the longest co-occurring sequence between a generated phrase and a reference phrase. BLEU-1, 2, and 3 are the cumulative 1-gram, 2-gram, and 3-gram scores for these pairs of phrases. 

Both BLEU and ROUGE results indicate that GPT-2 typically generates more accurate sentences for SAE than for AAVE (see Figures \ref{bleu-fig} and \ref{rouge-fig}). We note that the BLEU and ROUGE scores are relatively low since the comparison is between incomplete sentences of only five words.

We use a Wilcoxon rank-sum test to determine the significance of our BLEU and ROUGE results. With $\alpha = 0.05$, ROUGE-1 and ROUGE-L are significant. Additional p-values can be found in Table \ref{fig:bleu-rouge-sig-tests}.

\begin{table}[t!]
\begin{minipage}{0.47\textwidth}
\small
\centering
\setlength{\tabcolsep}{5.5pt}
\begin{tabular}{r|cccccc}
 & 
 \textsc{B-1} &
 \textsc{B-2} &
 \textsc{B-3} & 
 \textsc{R-1} &  
 \textsc{R-2} & 
 \textsc{R-L} \\
\hline
 \textit{p} & 0.256 & 0.095 & 0.097 & \textbf{0.001} & 0.811 & \textbf{0.003} \\
    \end{tabular}
        \caption{Wilcoxon rank-sum test p-values for each of our BLEU (B) and ROUGE (R) results. P-values that are significant with $\alpha = 0.05$ are in bold.}
    \label{fig:bleu-rouge-sig-tests}
\end{minipage}    
\vspace{-2ex}
\end{table}

\paragraph{Human Evaluation}

We also conduct human evaluation using AMT to assess the quality of the text generated by GPT-2. Annotators were filtered by HIT approval rate (higher than 95\%) and location (within the United States). They were given the first segment of an SAE phrase for context, followed by its corresponding GPT-2 generated segment. We did the same with each corresponding AAVE phrase. Annotators were asked to choose which one of the two generated phrases better fits the context of the respective first segment, which one has better quality, and which one is most likely machine-generated. Ties were allowed for this task. The annotator instructions for this task can be found in the \nameref{sec:human-eval-protocol}.

Results show that 21.7\% more annotators indicate that SAE generated segments have better quality than their corresponding AAVE generated segments, and 12\% more annotators indicate that SAE generated segments fit the context better than their AAVE generated segment counterparts (see Table \ref{fig:human-eval}). To determine existing bias in human evaluation, we perform the same evaluation on the original second segments of AAVE/SAE pairs and find that 48\% choose the SAE original second segments as likely machine-generated, while 31\% choose the AAVE original second segments. Looking at \ref{fig:human-eval}, the proportion of annotators who select SAE as machine generated decreases to 37.3\%, whereas the proportion for AAVE increases to 42.1\%. This indicates that GPT-2 worsens the quality of AAVE segments while improving the quality of SAE segments. These findings support our results from BLEU and ROUGE in demonstrating the unequal quality of GPT-2's text generation for SAE and AAVE, thus signifying a bias against AAVE.

\section{Conclusion}
Through this work, we highlight the need for AAVE-inclusivity in NLG models, especially those perceived as state-of-the-art. To this end, we provide a new evaluation of NLG models by comparing GPT-2's behavior on SAE and AAVE. In addition, we present a new dataset consisting of intent-parallel AAVE/SAE tweet pairs, which can be used in future works studying SAE and AAVE in NLP models. Our sentiment analysis experiments indicate that GPT-2 produces more negative instances when prompted with AAVE text. Moreover, our BLEU, ROUGE, and human evaluation results reveal a disparity in the quality of GPT-2's text generation between AAVE and SAE. We hope our findings can pave the way for further inclusion of diverse language in future NLG models.

\begin{table}[t!]
\begin{minipage}{0.47\textwidth}
\small
\centering
\setlength{\tabcolsep}{5.5pt}
\begin{tabular}{l|rrr}
 & \textit{Context} & \textit{Quality} & \textit{Likely MG} \\
\hline
 \textsc{SAE} & 48.7\% & 54.5\% & 37.3\% \\
 \textsc{AAVE} & 36.7\% & 32.8\% & 42.1\% \\
 \textsc{Tie}  & 14.6\% & 12.7\% & 20.6\% 
    \end{tabular}
        \caption{Human evaluation results, where ``MG" refers to ``Machine Generated." Tests are conducted pairwise between generated SAE and AAVE phrases.}
    \label{fig:human-eval}
\end{minipage}    
\vspace{-2ex}
\end{table}

\section*{Acknowledgements}

We thank our anonymous reviewers for their helpful feedback. This material is based upon work supported in part by the National Science Foundation under Grant 1821415. We would also like to thank the Amazon Alexa Knowledge team for their support. The authors are solely responsible for the contents of the paper, and the opinions expressed in this publication do not reflect those of the funding agencies.

\bibliography{anthology,emnlp2020}
\bibliographystyle{acl_natbib}

\newpage

\appendix

\section{Annotation Guidelines}
% \label{annotation-guidelines}

\subsection{AAVE to SAE protocol} \label{subsec:aave-to-sae}
You will be given a phrase that is written in African American Vernacular English, which we then split into two parts. Your task is to translate these parts one at a time into Standard American English so that your translations combine to form a coherent phrase.

Standard American English is used in a formal context, such as in professional communication. Although many of these phrases would not be used in professional communication, translate their vocabulary to SAE while maintaining their intent.

Specific cases:
\begin{enumerate}
	\item Acronyms: translate to its formal SAE equivalent (e.g. ``lol" $\rightarrow$ ``That's funny," ``I'm laughing,") or another equivalent for the given context. If the acronym expands to a valid SAE phrase, you can expand instead of providing a translation (e.g. ``ily" $\rightarrow$ ``I love you").
	\item Punctuation: translated phrases should have proper punctuation. Insert or fix capitalization, commas, periods, or other appropriate punctuation as necessary.
	\item Emoticons: remove emoticons from the translated phrase. For example, ``:)", ``:(", ``:/", and ``8)" should be removed.
    \item Phrase structure: translated phrases should maintain the structure as well as the intent of their original phrases. Keep general patterns, such as dependent or independent clauses. Try to keep the number of words in the translation about the same as the number of words in the original phrase.
    \item Translate the n-word to an appropriate equivalent.
    \item Keep swear words as is (the exception is the n-word. It needs to be translated as previously stated).
\end{enumerate}

\textbf{Example phrase:} So the hubby is out kickin it for him and his brutha bday so I guess i 'm going to bed alone tonight :(
\begin{itemize}
    \item \textbf{First part}: So the hubby is out kickin it for him and his brutha bday so I guess i 'm going
    \item \textbf{First part translated:} my husband is celebrating for his and his brother's birthday, so I guess I'm going
    \item \textbf{Second part:} to bed alone tonight :(
    \item \textbf{Second part translated:} to bed alone tonight.
\end{itemize}

\subsection{Semantic equivalence protocol} \label{subsec:semantic-equivalence-guidelines}
Given two phrases, determine whether or not their semantics (meanings) are the same.
\begin{itemize}
    \item Are you a native or fluent AAVE speaker?
    \item If you responded `Yes' above, are you able to code switch?
    \item Are Phrase 1 and Phrase 2 semantically equivalent?
\end{itemize}

\subsection{Human evaluation protocol}
\label{sec:human-eval-protocol}
You will be given two long phrases. Each one will be split into two parts. Your task is to evaluate the quality of the second part of each phrase and determine which one you prefer.

Looking at the second part of both phrases:
\begin{itemize}
    \item Which one has better quality?
    \item Which one fits the context of its respective first part better?
    \item Which one is more likely generated by machines?
\end{itemize}

\end{document}